\documentclass[twoside,11pt]{article}

\usepackage[abbrvbib]{jmlr2e}

\usepackage[utf8]{inputenc} 
\usepackage[T1]{fontenc}
\usepackage{tikz}
\usepackage{wrapfig}
\usepackage{caption,subcaption}
\usepackage{multirow}
\usepackage{pbox}
\usepackage{colortbl}
\usepackage{hyperref}
\usepackage{makecell}
\usepackage{listings}
\usepackage{hyperref}       
\usepackage{url}            
\usepackage{booktabs}       
\usepackage{amsfonts}       
\usepackage{nicefrac}       
\usepackage{microtype}      
\usepackage{graphicx}
\usepackage{graphics}
\usepackage{xcolor}
\usepackage{soul}
\usepackage[frozencache=true,cachedir=minted-cache-1]{minted}
\usepackage{lscape}
\usepackage{multirow}
\usepackage{longtable}
\usepackage{here}
\usepackage{threeparttable}
\definecolor{LightGray}{gray}{0.9}
\hypersetup{
  colorlinks=true,
  citecolor=black,
  linkcolor=black,
  urlcolor=black}

\usepackage{lastpage}
\jmlrheading{24}{2023}{1-\pageref{LastPage}}{3/23; Revised
7/23}{8/23}{23-0378}{Siyi Hu, Yifan Zhong, Minquan Gao, Weixun Wang, Hao Dong, Zhihui Li, Xiaodan Liang, Xiaojun Chang and Yaodong Yang}
\ShortHeadings{MARLlib: A Scalable and Efficient Multi-agent Reinforcement Learning Library}{Hu, Zhong, Gao, Wang, Dong, Li, Liang, Chang and Yang}

\firstpageno{1}

\begin{document}

\title{MARLlib: A Scalable and Efficient Library For Multi-agent Reinforcement Learning }

\author{\name Siyi Hu$^{1}$ \email siyi.hu@student.uts.edu.au\\\qquad
\name Yifan Zhong$^{2}$ \email zhongyifan@stu.pku.edu.cn \\\qquad
\name Minquan Gao$^{2}$ \email minchiuan.gao@gmail.com\\\qquad
\name Weixun Wang$^{3}$ \email wangweixun@corp.netease.com \\\qquad
\name Hao Dong$^{2}$ \email hao.dong@pku.edu.cn \\\qquad
\name Xiaodan Liang$^{4,6}$ \email xiaodan.liang@sysu.edu.cn\\\qquad
\name Zhihui Li$^{5}$ \email zhihuilics@gmail.com\\\qquad
\name Xiaojun Chang$^{1,4} \dagger$ \email xiaojun.chang@uts.edu.au\\\qquad
\name Yaodong Yang$^{2} \dagger$ \email yaodong.yang@pku.edu.cn\\\qquad
\addr 
$^{1}$ ReLER, AAII, University of Technology Sydney\\
$^{2}$ Institute for Artificial Intelligence, Peking University\\
$^{3}$ NetEase Fuxi AI Lab \quad $^{4}$ MBZUAI \\
$^{5}$ Shandong Artificial Intelligence Institute, Qilu University of Technology  \\ 
$^{6}$ School of Intelligent Systems Engineering, Sun Yat-sen University  \\
$\dagger$ corresponding authors
\AND
}
\editor{Zeyi Wen}

\maketitle

\begin{abstract}
A significant challenge facing researchers in the area of multi-agent reinforcement learning (MARL) pertains to the identification of a library that can offer fast and compatible development for multi-agent tasks and algorithm combinations, while obviating the need to consider compatibility issues. In this paper, we present MARLlib, a library designed to address the aforementioned challenge by leveraging three key mechanisms: 1) a standardized multi-agent environment wrapper, 2) an agent-level algorithm implementation, and 3) a flexible policy mapping strategy. By utilizing these mechanisms, MARLlib can effectively disentangle the intertwined nature of the multi-agent task and the learning process of the algorithm, with the ability to automatically alter the training strategy based on the current task's attributes. The MARLlib library's source code is publicly accessible on GitHub: \url{https://github.com/Replicable-MARL/MARLlib}.
\end{abstract}

\vspace{-10pt}
\section{Introduction}
\vspace{-5pt}
The field of Multi-Agent Reinforcement Learning (MARL) has garnered significant attention for its real-world applications and potential in enhancing collective intelligence \citep{busoniu2008comprehensive, yang2020overview, zhang2021multi}. Earlier works have shown that agents can acquire strategies that surpass the capabilities of human experts and facilitate human decision-making in a retrograde fashion \citep{vinyals2019grandmaster, baker2019emergent}.

While notable advancements have been made in single-agent reinforcement learning \citep{baselines, liang2018rllib, weng2021tianshou}, the development of a comprehensive and high-quality MARL library presents distinct challenges.
One major challenge arises from the absence of a standard dataset for evaluating new ideas, unlike domains such as image classification that utilize representative datasets like ImageNet \citep{deng2009imagenet}. MARL datasets encompass customizable scenarios with varying agent numbers, map sizes, reward functions, and unit statuses, making it difficult to establish a research starting point. Efforts have been made to create unified and scalable testing suites for the MARL community, such as PettingZoo \citep{terry2020pettingzoo} and Melting Pot \citep{leibo2021meltingpot}. However, practical challenges persist, including variations in lower-level data structures, requiring adjustments to the learning pipeline which impacts algorithm performance and reliability.

\begin{table*}[t]
		\caption{A comparison between current MARL libraries and our MARLlib.}
		\label{table:marl benchmark}
        \vspace{-20pt}
		\begin{center}
			\resizebox{1.\textwidth}{!}{
				\begin{tabular}{c c c c c c}
					\toprule
					\textbf{Library} & \pbox{50pt}{\centering\textbf{Supported Env}} & \pbox{60pt}{\centering\textbf{Algorithm}} & \pbox{60pt}{\centering\textbf{Parameter Sharing}} &  \pbox{70pt}{\centering\textbf{Agent Architecture}}  & \pbox{90pt}{\centering\textbf{Framework}} 
					\\  \midrule \pbox{80pt}{\centering \href{https://github.com/oxwhirl/pymarl}{PyMARL}} & \makecell{ 1\\cooperative}  & \makecell{5} & share & GRU & PyMARL
					\\ \arrayrulecolor{black!30}\midrule 
					\pbox{80pt}{\centering \href{https://github.com/hijkzzz/pymarl2}{PyMARL2}}  & \makecell{2\\cooperative}  &\makecell{11} & share  & MLP+GRU &  PyMARL
					\\ \arrayrulecolor{black!30}\midrule 
					\pbox{80pt}{\centering \href{https://github.com/starry-sky6688/MARL-Algorithms}{MARL-Algorithms}} & \makecell{1\\cooperative} & \makecell{9} &  share & MLP+GRU  & MARL-Algorithms
					\\ \arrayrulecolor{black!30}\midrule 
					\pbox{80pt}{\centering \href{https://github.com/uoe-agents/epymarl}{EPyMARL}} & \makecell{4\\cooperative}  &   \makecell{9}  & \makecell{share + separate}   & GRU &  PyMARL
					\\ \arrayrulecolor{black!30}\midrule 
					\pbox{80pt}{\centering \href{https://github.com/sjtu-marl/malib}{MAlib}} & \makecell{4\\self-play} &   \pbox{90pt}{9}   & \makecell{share+group+separate}  & MLP+LSTM & MAlib
					\\ \arrayrulecolor{black!30}\midrule 
					\pbox{80pt}{\centering \href{https://github.com/marlbenchmark/on-policy}{MAPPO}} &  \makecell{4\\cooperative} &  1 & \makecell{share + separate}  & MLP+GRU+CNN  & \makecell{pytorch-a2c-ppo-acktr\\-gail \citep{pytorch-a2c-ppo-acktr-gail}}
					\\ \arrayrulecolor{black!30}\midrule \midrule 
					\pbox{80pt}{\centering \href{https://github.com/Replicable-MARL/MARLlib}{MARLlib}} & \makecell{15+\\ \bf{no restriction}} &  \makecell{ \bf{18}} & \makecell{share+group+ separate \\ \bf{customizable}}  & \makecell{MLP+LSTM+GRU \\ +CNN} & \makecell{Ray \citep{moritz2018ray} \\ RLlib \citep{liang2018rllib}}
					\\ \arrayrulecolor{black} 
					\bottomrule
				\end{tabular}
                
			}
		\end{center}
  \vspace{-25pt}
	\end{table*}

Another significant challenge stems from the inherent incompatibility between multi-agent environments and algorithms. For instance, the coexistence of cooperative and competitive learning targets within a single environment hinders the direct application of algorithms designed solely for cooperative use. Additionally, specific algorithms may require additional task information, such as global state, rendering them incompatible with environments lacking such data. Because of these, existing libraries \citep{papoudakis2021benchmarking, hu2021rethinking, yu2021surprising} suffer from limited task coverage and lack algorithm unification, resulting in poor extensibility and a bloated code structure (see Table~\ref{table:marl benchmark}).

Therefore, the development of a universal learning framework that effectively disentangles the environment and the algorithm, ensures compatibility, and provides a standard testing suite for MARL is crucial. In this paper, we introduce MARLlib, a new library that combines the core advantages of Ray \citep{moritz2018ray} and RLlib \citep{liang2018rllib}, while incorporating novel features. These features include a standardized multi-agent environment wrapper, agent-level algorithm implementation, and a flexible policy mapping strategy. MARLlib serves as a scalable and efficient framework for the MARL research community, enabling the construction, training, and evaluation of MARL algorithms across diverse multi-agent environments.

\vspace{-10pt}
\section{Design of MARLlib}
\vspace{-2pt}
\label{sec:design}

 	\begin{figure*}[t]
        \vspace{-15pt}
		\centering 
			\includegraphics[width=\linewidth]{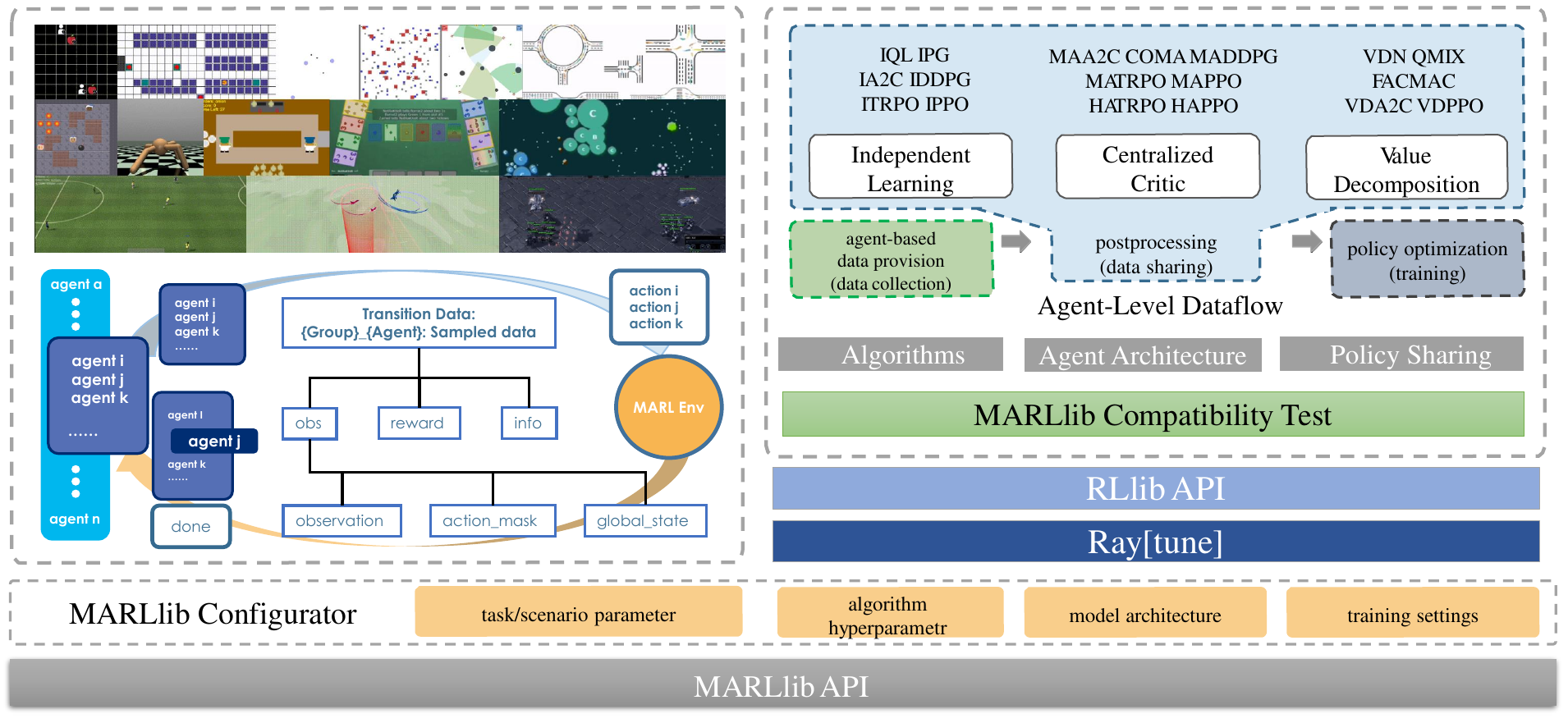}
		\caption{Overview of MARLlib, a framework that integrates algorithms and tasks into a unified framework.}
		\label{fig:marllib_open}
		\vspace{-20pt}
	\end{figure*}

MARLlib's architecture is comprised of three key components: 1) a standardized wrapper which is utilized to unify the multi-agent environment; 2) the algorithm implemented at the agent-level dataflow; 3) a flexible policy mapping strategy to address compatibility issues that may arise between tasks and the multi-agent learning process.

\textbf{Multi-agent environment wrapper.}
A standardized agent-environment interaction style, such as Gym \citep{openaigym}, can effectively separate the environment from the algorithm, thereby allowing the learning process to remain agnostic to the tasks. However, extending this approach to multi-agent settings is not straightforward. Specifically, there are two primary issues that must be addressed. Firstly, the structure of the data returned by multi-agent environments can vary significantly, posing a challenge for the algorithm-side learning pipeline's design. Secondly, given the need for agents to act simultaneously or asynchronously, a flexible data collection process is essential. In light of this, the development of a new environment wrapper for multi-agent environment is deemed a pressing matter.

To tackle the first issue, it is necessary to consider the common data provision in multi-agent tasks. Upon examining their data structure, we identify the following significant differences: 1) the provision of data, where information that is not universally required may be missed (such as global state and action mask), 2) the format in which rewards are provided, either as a scalar (e.g., team reward) or a set (with each agent receiving its own reward), and 3) whether each agent's data is labeled with an agent ID or centrally collected without labeling. In order to reconcile these differences, we propose a general environment wrapper that can effectively mitigate discrepancies between multi-agent tasks and align them with a similar data provision style. This entails three crucial steps: 1) the inclusion of non-universal data, such as global state, into the observation, 2) the duplication and allocation of rewards to each agent to ensure equity, and 3) the labeling of all collected data with the agent ID, enabling each agent to manage its own data and uphold parity with other agents.

Towards the second issue, we leverage the advanced data collection mechanism provided by RLlib. In this approach, each agent collects data independently and maintains the sampled data in its individual buffer. Agents are not required to complete the data sampling process at the same time stamp. One episode sampling process is ended when the \texttt{done} signal transitions to \texttt{True}, indicating the completion of the sampling process of all agents. By addressing these two issues, MARLlib's task-side has been purposefully crafted to enable flexible but standardized data collection, thereby providing the algorithm-side with a clear and consistent source data style. By integrating this approach, MARLlib empowers the corresponding algorithm-side to effectively leverage the benefits offered by the task-side, thereby enabling scalable multi-agent learning.

\textbf{Agent-level dataflow.}
The new environment wrapper unifies data provision style at the agent level, and it would be convenient if the algorithm side could also be implemented in this way to allow for seamless data flow. The key question is whether MARL algorithms can be transformed into an agent-level learning process. In MARLlib, we primarily focus on decentralized partially observable Markov decision process (Dec-POMDP) settings \citep{oliehoek2016concise} and three typical learning styles: independent learning (IL), centralized critic (CC), and value decomposition (VD). Other algorithms can be extended from these to fulfill various training requirements.
Figure \ref{fig:algorithm} provides an illustration of the agent-level perspective of the MARL process. For IL, the learning process can be naturally decomposed into agent-level dataflow, as one agent treats other agents as part of the environment and thus maintains its independency. For CC, sharing information in the post-processing phase to form the centralized input can transform the CC pipeline into agent-level dataflow. For VD, exchanging predicted Q/critic value before entering the training stage enables mixed value functions to work with a single agent, thereby transforming the VD pipeline into agent-level dataflow. More detailed discussion of the equivalence between central and agent-level dataflow can be found in Appendix.~\ref{equivalence}.

\textbf{Policy mapping strategy.}
Upon establishing both the agent-based data provision on the environment side and the agent-level dataflow on the algorithm side, it becomes imperative to devise a mechanism for managing the dataflows between them that ensures the appropriate allocation of agents to their respective learning targets. In this regard, we leverage RLlib's policy mapping API, enrich it, and standardize strategy construction into three steps: (i) provision of a policy mapping dictionary by the environment side to describe task abstraction, agent group, and restrictions on parameter sharing; (ii) selection of a suitable strategy from default settings, including full-sharing, non-sharing, and group-sharing, or customization of one by users; and (iii) verification by the algorithm side of the legality of the chosen policy mapping strategy.

Other noteworthy attributes encompass automatic model construction, a thoroughly decoupled configuration framework, conveniently accessible plug-in/out strategies, and an automated compatibility assessment. Notably, MARLlib is accompanied by a comprehensive \href{https://marllib.readthedocs.io/}{\textit{documentation}} that encompasses four distinct segments: the MARLlib handbook, guidance tailored for newcomers to the MARL domain, algorithmic documentation, and an extensive survey of MARL methodologies. 
Moreover, a diverse spectrum of \href{https://github.com/Replicable-MARL/MARLlib/tree/master/results}{\textit{experimental endeavors}} has been undertaken employing MARLlib.

\label{sec:benchmark}

\vspace{-5pt}
\section{Conclusion}
\vspace{-3pt}

In this paper, we present MARLlib, a unified library suite that encompasses a vast array of algorithms and tasks in the MARL domain. This comprehensive library suite is designed to provide a dependable and comprehensive toolset for training, evaluating, and comparing MARL algorithms. The MARL community can take advantage of MARLlib to build a diverse range of multi-agent applications. Furthermore, MARLlib can serve as an educational platform for new researchers and contribute to the growth of the MARL research field.

\vskip 0.2in


\newpage
\appendix

\newpage
\section*{Acknowledgement}
This work is sponsored by National Key R\&D Program of China (2022ZD0114900), Beijing Municipal Science \& Technology Commission (Z221100003422004), Young Elite Scientists Sponsorship Program by CAST (2022QNRC002), Outstanding Youth Fund of Shandong Province (ZR2021YQ44), and ``Taishan Scholars Youth Expert Program'' of Shandong Province.

\section*{Appendices}
\section{Installation}

The installation of MARLlib has two parts: common installation and external environment installation. We have tested the installation on Python 3.8 with both Ubuntu 18.04 and Ubuntu 20.04.

 \subsection{Install dependencies (basic)}

We strongly recommend using conda to manage your dependencies and avoid version conflicts. Here's an example of building a Python 3.8 based conda environment.

     \begin{minted}
    [
    frame=lines,
    bgcolor=LightGray,
    breaklines,
    fontsize=\scriptsize
    ]{bash}
    
    $ conda create -n marllib python=3.8
    $ conda activate marllib
    $ git clone https://github.com/Replicable-MARL/MARLlib.git
    $ cd MARLlib
    $ pip install --upgrade pip
    $ pip install -r requirements.txt
    
    # recommend always keeping the gym version at 0.21.0.
    $ pip install gym==0.21.0
    
    # add patch files to MARLlib
    $ python patch/add_patch.py -y
    \end{minted}

 \subsection{Install environments (optional)}

External environments are not auto-integrated. However, you can install them by following this \href{https://marllib.readthedocs.io/en/latest/handbook/env.html}{link}. We recommend always keeping the gym version at 0.21.0, which is the compatible version for all integrated tasks.

\vspace{-5pt}
\section{Library usage}
\vspace{-10pt}
\begin{minted}
[
frame=lines,
bgcolor=LightGray,
breaklines,
fontsize=\scriptsize
]{python}
from marllib import marl

# prepare env
env = marl.make_env(environment_name="mpe", map_name="simple_spread")
# initialize algorithm with appointed hyper-parameters
mappo = marl.algos.mappo(hyperparam_source="mpe")
# build agent model based on env + algorithms + user preference
model = marl.build_model(env, mappo, {"core_arch": "mlp", "encode_layer": "128-256"})
# start training
mappo.fit(
  env, model, 
  stop={"timesteps_total": 1000000}, 
  checkpoint_freq=100, 
  share_policy="group"
)
# rendering
mappo.render(
  env, model, 
  local_mode=True, 
  restore_path={'params_path': "checkpoint_000010/params.json",
                'model_path': "checkpoint_000010/checkpoint-10"}
)
\end{minted}
MARLlib offers an API that provides a user-friendly programming interface, simplifying the utilization of the library, while still maintaining a high degree of extensibility to facilitate user customization. This allows researchers to concentrate on their specific research goals without being bogged down by implementation details.

\section{Training Efficiency}

We conducted experiments to demonstrate the efficiency of MARLlib compared to EPyMARL and the on-policy baseline (official MAPPO \citep{yu2021surprising}). The experiments were performed on a local server with an NVIDIA RTX A6000 GPU and an AMD Ryzen Threadripper PRO 5945WX 12-Cores CPU. The testing scenario is MMM2 from SMAC \citep{samvelyan2019starcraft}, and the testing algorithm is MAPPO. The total consumed timesteps are $10^6$.

From Table.~\ref{table:effiency}, it is evident that MARLlib is significantly more efficient than the other frameworks in terms of clock time. However, the increased training speed comes with relatively higher memory usage and GPU memory usage. This is partly due to Ray/RLlib's scheduling mechanism, where data is cached on the GPU as long as there is still available memory. Additionally, since each agent is asked to maintain its own data buffer in MARLlib (which is a prerequisite for tackling task modes such as competitive and mixed), the memory usage is higher than in the other two frameworks.
 
    \begin{table*}[!h]
		\centering
		\vspace{-1pt}
		\caption{Efficency and hardware usage comparison between EPyMARL, MAPPO, and MARLlib}
		\vspace{-1pt}
    \begin{threeparttable}
		\begin{tabular}{c|ccc}
                \toprule
			Framework & Clocktime (min:sec) & Memory (GB) & GPU Memory (MB) \\ 
                \midrule
			EPyMARL(thread=5/10/15) & 5:29/3:14/2:24 & 8.4/12.3/15.8 & 2245/2309/2329 \\ 
			MAPPO(thread=5/10/15) & 5:12/3:13/2:42 & 8.9/12.3/16.3 & 2157/2277/2389 \\ 
			MARLlib(worker=5/10/15) & 3:29/2:16/1:24 & 11.2/15.4/20.4 & 5025/5327/5351 \\
                \bottomrule
		\end{tabular}
    \end{threeparttable}
    \label{table:effiency}
	\end{table*}

\section{Parameter}

In MARLLib, there are three types of hyperparameters to choose from when initializing algorithms: \texttt{common}, \texttt{finetuned}, and \texttt{test}. The \texttt{common} hyperparameters are suitable for normal training on new tasks where the optimal hyperparameters are unknown. For a fair comparison with other algorithms on commonly used MARL tasks like MPE and SMAC, we recommend using the \texttt{finetuned} hyperparameters. The \texttt{test} hyperparameters are intended for developing and testing new algorithms or incorporating new multi-agent environments. You can find all the available parameters by visiting this \href{https://github.com/Replicable-MARL/MARLlib/tree/master/marllib/marl/algos/hyperparams}{link}.

\section{Connection to RLlib}

MARLlib and RLlib are closely related, with MARLlib building upon the foundation provided by RLlib. MARLlib extends and enhances RLlib's capabilities specifically in the domain of multi-agent reinforcement learning (MARL). It leverages RLlib's infrastructure, including its multi-agent task interface, to create a unified and compatible agent-environment interface for MARL experiments. This relationship allows researchers and developers to benefit from both the versatility and functionality of RLlib while harnessing the specialized features and optimizations provided by MARLlib for MARL tasks.

In the following section, we will discuss the challenges of multi-agent learning in RLlib and the major improvements proposed by MARLlib that make it stand out as a comprehensive library, not limited to being just a "plug and play" extension for RLlib.

\subsection{Challenges of RLlib's Multi-Agent Case}

While RLlib provides a robust infrastructure for reinforcement learning, the multi-agent case within RLlib poses certain challenges that can make it difficult to use effectively. These challenges stem from:

\begin{itemize}
    \item \textbf{Lack of a Standardized and Unified Agent-Environment Interface for Multi-Agent Tasks}: Multi-agent reinforcement learning involves multiple agents interacting with an environment, which introduces complexity in designing the agent-environment interface. However, RLlib's multi-agent case suffers from the lack of a standardized and unified interface. The absence of clear conventions for representing and exchanging information between agents and the environment hinders the development and comparison of different MARL algorithms.
    \item \textbf{Complexity and Accessibility for Newcomers}: The multi-agent case\footnote{An example can be found in this link: \url{https://github.com/ray-project/ray/blob/master/rllib/examples/multi_agent_custom_policy.py}} in RLlib often requires deep knowledge of the underlying RLlib framework, making it less accessible to newcomers in the field of multi-agent reinforcement learning. Effectively utilizing RLlib's multi-agent functionality necessitates prior understanding of advanced RL concepts, which can act as a barrier for researchers and practitioners who are new to multi-agent RL.
    \item \textbf{Lack of a Focused Point for Integrating Different Algorithms}: RLlib's multi-agent case is built without a focused point of integration for different algorithms. This lack of a unifying framework makes it challenging to compare and combine different MARL algorithms within RLlib effectively. Researchers and practitioners may find it difficult to implement and evaluate their own algorithms in a standardized manner, hindering the progress and collaboration in the field.
\end{itemize}

Addressing these challenges is crucial for advancing multi-agent reinforcement learning research. Efforts to simplify and improve the usability of RLlib's multi-agent case to provide researchers and developers with a more accessible and efficient framework for conducting multi-agent experiments need to be made.

\subsection{MARLlib vs RLlib}

MARLlib builds upon RLlib's foundation and addresses the shortcomings of RLlib's multi-agent case, offering significant improvements in the field of multi-agent reinforcement learning.

\begin{table}[htbp]
  \centering
  \caption{Comparison of MARLlib and RLlib on multi-agent case}
  \label{tab:marllib-vs-rllib}
\begin{threeparttable}
  \begin{tabular}{lcc}
    \toprule
    \textbf{Features} & \textbf{MARLlib} & \textbf{RLlib} \\
    \midrule
    Task Interface \& Transition Data & Structured & Vague \& Flexible \\
    \hline
    Multi-agent Algorithm Support & Standard CTDE & \makecell{Simple Extension on  RL} \\
    \hline
    Policy Mapping \& Sharing & Auto & Manual \\
    \hline
    Scalability and Interoperability & Inherited & \checkmark \\
    \hline
    Accessibility for newcomers & Easy & Hard \\
    \hline
    Auto Adaption \& Compatibility Test  & \checkmark &  $\times$\\
    \hline
    Experimentation \& Benchmarking & \checkmark & Limited \\
    \hline
    Documentation and tutorials & Comprehensive & Limited \\
    \bottomrule
  \end{tabular}
  \end{threeparttable}

\end{table}



The connection between MARLlib and RLlib mirrors the integration and functionality similarity observed in TensorFlow \citep{abadi2016tensorflow} and Keras \citep{chollet2015keras}. While Keras serves as a high-level neural networks API with TensorFlow as its efficient backend for execution and optimization, MARLlib leverages RLlib's reinforcement learning infrastructure for effective development of multi-agent systems through its user-friendly API. These connections facilitate streamlined development, abstracting complexities while benefiting from powerful machine learning and reinforcement learning capabilities.

MARLlib's primary contribution to RLlib is providing a unified and compatible agent-environment interface, simplifying data consumption for multi-agent algorithms across a wide range of tasks. This improvement promotes efficiency and ease of implementation in multi-agent reinforcement learning research.

Secondly, MARLlib leverages RLlib's agent-side abstraction and categorization to address the complexities inherent in diverse multi-agent tasks. By renovating the information sharing stage, MARLlib ensures effective sharing and compatibility across algorithms, aiming to unify algorithms under one framework. It builds upon RLlib's reliable implementation of single-agent reinforcement learning, requiring only necessary adjustments for multi-agent scenarios.

MARLlib's third contribution to RLlib lies in its user-friendly approach, simplifying the learning and exploration of multi-agent reinforcement learning for newcomers. With a simplified API, MARLlib handles the compatibility between diverse multi-agent tasks and algorithms, allowing users to focus on core components, such as the selection of algorithms, tasks, and models. However, this ease of use does not limit the extension or modification of MARLlib's learning pipeline. Users can introduce new environments \footnote{\url{https://github.com/Replicable-MARL/MARLlib/blob/master/examples/add_new_env.py}}, customize policy sharing \footnote{\url{https://github.com/Replicable-MARL/MARLlib/blob/master/examples/customize_policy_sharing.py}}, and modify the information sharing stage to define different aspects of observation sharing. These features empower users to tailor their own experiments based on the flexibility provided by MARLlib.

In addition, MARLlib offers users novel features, including pipeline auto-adaptation and training compatibility testing, complemented by comprehensive documentation on multi-agent reinforcement learning.

\section{Related Work}

In the realm of machine learning, it is widely acknowledged that evaluating a novel idea necessitates the use of an appropriate dataset. The prevailing approach entails employing a representative or widely accepted dataset, adhering to its established evaluation pipeline, and comparing the performance of the new idea against existing algorithms. A prominent example is the ImageNet dataset (Deng et al., 2009), which has served as the benchmark for image classification, a fundamental task in computer vision, for nearly a decade since the advent of deep learning.

However, this paradigm does not readily apply to Multi-Agent Reinforcement Learning (MARL). In the context of MARL, a dataset corresponds to a collection of scenarios that constitute a multi-agent task. Multi-agent tasks exhibit a high degree of customization, encompassing various aspects such as the number of agents, map size, reward function, and unit status. The range of possible adjustments to multi-agent tasks or environments is vast, with new tasks and environments constantly emerging. Consequently, for newcomers in the field, selecting an appropriate starting point for MARL research assumes paramount importance: one must first identify a set of widely used multi-agent datasets and then strive to make improvements upon existing ones. This approach has become prevalent in recent MARL publications, employing standardized benchmarking processes and testing suites such as MPE, SMAC, and MAMuJoCo.

However, it is important to note that the superiority of an algorithm should not solely be measured by its performance in specific tasks. In reality, there exists an ongoing debate regarding whether MARL algorithms should focus on excelling in a single task or aim for strong performance across multiple tasks, presenting a dilemma commonly known as the "algorithm-first or task-first" conundrum. MARLlib leans towards the "algorithm-first" perspective, as we believe that it is relatively easier for other users to reproduce and compare algorithms in this manner, fostering a trustworthy solution that propels the advancement of the field.

Following the "algorithm-first" principle, there is an urgent need to construct a comprehensive collection of standardized multi-agent tasks, accompanied by a unified algorithmic learning pipeline. Here, we delve into the contributions made by the MARL community from two perspectives: the task side and the algorithm side.


On the task side, several notable endeavors, such as PettingZoo \citep{terry2020pettingzoo} and Melting Pot \citep{leibo2021meltingpot}, have aimed to create a unified and scalable testing suite for the MARL community. However, practical challenges persist. For instance, in PettingZoo, the interface appears to be unified at a high level (agent-env interface), but exhibits variations at a lower level (inside data structure). Consequently, directly feeding the returned data into the learning pipeline is not suitable. Adjustments must be made to the learning pipeline to accommodate the unique characteristics of each task. The manner in which these adjustments are implemented plays a pivotal role in the performance of the algorithm, thus significantly impacting the reliability of the algorithm's learning curve. Another significant challenge arises from the incompatibility between interfaces like PettingZoo and popular MARL tasks such as SMAC \citep{samvelyan2019starcraft}, a lighter version of the full game StarCraft II \citep{silver2017mastering}. This incompatibility poses difficulties when evaluating algorithms across different testing suite.

Nevertheless, there are existing libraries that support multiple tasks, often by forgoing the unification of task-side interfaces. For instance, the official MAPPO implementation creates a separate runner for each distinct task. As of now, the MAPPO benchmark supports four environments and has accordingly prepared more than six runner files including shared and non-shared styles \footnote{\url{https://github.com/marlbenchmark/on-policy/tree/main/onpolicy/runner}}. However, this approach is not without limitations, as new tasks continually emerge, making the maintenance of a large number of runners impractical.

\label{appx:framework_compare}

After the discussion on the current progress on the task side, we will now shift our focus to the algorithm side. Instead of conducting an algorithm-level comparison to determine which algorithm is better, our aim is to explore previous efforts in incorporating as many algorithms as possible into a single framework. We present a comparison table that showcases some of these frameworks in Table~\ref{table:marl benchmark}. Here, we provide a brief introduction to each of them:
 
\begin{itemize}

	\item {\bf \href{https://github.com/oxwhirl/pymarl}{PyMARL}} \citep{samvelyan2019starcraft}  is the first and most well-known MARL library. All algorithms in PyMARL are built for SMAC \citep{samvelyan2019starcraft}, where agents learn to cooperate for a higher team reward. However, PyMARL has not been updated for a long time and can not catch up with the recent progress. To address this, the extension versions of PyMARL are presented including PyMARL2 \citep{hu2021rethinking} and EPyMARL \citep{papoudakis2021benchmarking}.
	
	\item {\bf \href{https://github.com/hijkzzz/pymarl2}{PyMARL2}} 
     \citep{hu2021rethinking} is an extension of PyMARL and still focuses on credit assignment mechanism. It provides a finetuned QMIX \citep{rashid2018qmix} with state-of-art-performance on SMAC. The number of available algorithms increases to ten, with more code-level tricks incorporated.

     \item {\bf \href{https://github.com/starry-sky6688/MARL-Algorithms}{MARL-Algorithms}} \citep{MARLAlgorithms} is a library that covers broader topics compared to PyMARL including learning better credit assignment, communication-based learning, graph-based learning, and multi-task curriculum learning. Each topic has at least one algorithm, with nine implemented algorithms in total. The testing bed is limited to SMAC, a cooperative mult-agent benchmark.
  
	\item {\bf \href{https://github.com/uoe-agents/epymarl}{EPyMARL}} \citep{papoudakis2021benchmarking} is another extension for PyMARL that aims to present a comprehensive view on how to unify cooperative MARL algorithms. It first proposed independent learning, value decomposition, and centralized critic categorization but is restricted to cooperative settings. Nine algorithms are implemented in EPyMARL. Three more cooperative environments LBF \citep{christianos2020shared}, RWARE \citep{christianos2020shared}, and MPE \citep{lowe2017multi} are incorporated to evaluate the algorithms on cooperative MARL.
	
	\item {\bf \href{https://github.com/marlbenchmark/on-policy}{MAPPO benchmark}} \citep{yu2021surprising} is the official code base of MAPPO \citep{yu2021surprising}. It focuses on cooperative MARL and covers four environments including SMAC, MPE, Hanabi \citep{bard2020hanabi}, and Google Research Football \citep{kurach2019google}. It aims at building a strong baseline and only contains MAPPO.
	
	\item {\bf \href{https://github.com/sjtu-marl/malib}{MAlib}} \citep{zhou2021malib} is a recent library for population-based MARL which combines game-theory and MARL to solve multi-agent tasks in the scope of meta-game.
 
\end{itemize}

 	Existing libraries and benchmarks provide good platforms for developing and comparing MARL algorithms in different environments. However, there are essential limitations to the current work. Firstly, these works are limited in task coverage due to the lack of a unified agent-env interface, as mentioned earlier. Moreover, existing work pays little attention to how algorithms are organized, focusing primarily on implementing existing algorithms. This results in poor extensibility and a bloated code structure in the algorithm-side learning pipeline.

\section{MARLlib's Solution on Task Side Unification}
\label{sec: task-side uni}
A crucial lesson learned from the community is that unification must be based on establishing a standardized task suite interface that standardizes the sampling stage while ensuring compatibility with the underlying nature of different tasks and their corresponding data structures.

MARLlib's agent-env interface, as highlighted in the main paper, has successfully enabled seamless integration of diverse multi-agent tasks and types of MARL algorithms. In the subsequent sections, we present an in-depth exploration of MARLlib's interface capabilities, showcasing its ability to handle the intricate demands of different tasks and data formats while ensuring data coherence.


\subsection{Support for Task Characteristics}

MARLlib's interface accommodates a wide range of task characteristics, including various learning modes (cooperative, collaborative, competitive, mixed), observability levels (full, partial), action spaces (discrete, continuous, multi-discrete), observation space dimensions (1D, 2D), action masking, the presence of a global state, and different types of rewards (dense, sparse). By encompassing these diverse characteristics, MARLlib provides a flexible framework for conducting experiments on a broad spectrum of multi-agent tasks. Importantly, the task characteristics are automatically detected by the algorithm learning pipeline, eliminating the need for manual attention to compatibility. After a successful training launch, errors are triggered if there is an incompatible combination, such as using VDA2C \citep{su2021value} (cooperative only) with the \textit{simple\_adversary} scenario in MAgent, based on MARLlib's compatibility test.

\begin{table}[h]
\centering
\begin{tabular}{ll}
\hline
\textbf{Learning Mode} & Cooperative + Collaborative + Competitive + Mixed \\
\textbf{Observability} & Full + Partial \\
\textbf{Action Space} & Discrete + Continuous + MultiDiscrete \\
\textbf{Observation Space Dim} & 1D - 3D \\
\textbf{Action Mask} & Yes \\
\textbf{Global State} & Yes \\
\textbf{Reward} & Dense + Sparse \\
\textbf{Agent-Env Interact Mode} & Simultaneous + Asynchronous \\
\hline
\end{tabular}
\end{table}

\subsection{Data Inner Alignment}

The data sampling process is standardized in existing multi-agent task suites such as PettingZoo, but only on the code level. By saying "code level," we mean that under the multi-agent setting, the data returned by the agent-environment interface may not actually match what the variable name suggests. For example, observations containing global information should never be fed directly into actor networks of algorithms like MAPPO or VDA2C, as this would violate the principles of centralized training and decentralized execution (CDTE) settings.

To address this challenge, MARLlib ensures strict data alignment in all incorporated environments. This alignment guarantees that the data returned from the environment can be properly utilized by the algorithms on the algorithm side. For instance, in many tasks, a natural global state may not be readily available. However, the global information can be extracted from raw observations (e.g., MAgent) or by concatenating the observations of all agents into a single global observation (e.g., MPE). This adoption is automatically accomplished by MARLlib.

\section{MARLlib's Solution on Algorithm Side Unification}

MARLlib effectively solves the dilemma by providing better algorithm unification and categorization, implementing more algorithms both in quantity and diversity, with no task mode restriction, supporting 15 environment suites, and allowing flexible parameter sharing. The core idea of all these features is agent-level dataflow, which is discussed in the main paper. There are two more main features that need to be clarified:

The first feature is how MARLlib manages to decompose the complexity of MARL algorithms and step forward to an all-task-mode framework that no existing MARL library can achieve. From the existing libraries listed in the previous section, a major challenge arises in breaking the restriction on applicable task modes. Most libraries initially focus on cooperative tasks, making it technically impossible to incorporate new task modes such as competitive tasks (e.g., Pommerman \citep{pommerman}) and mixed tasks (e.g., MAgent \citep{zheng2017magent}). In MARLlib, the agent-level dataflow addresses this challenge by treating each agent's learning process separately from others. This means that the learning target can be decomposed to the individual level, enabling agents to learn in any task mode and style (collectively, in groups, or individually). However, this is not the ultimate solution, as the nature of different algorithm types limits their application. Several commonly observed cases are discussed below:

\begin{itemize}
    \item Value Decomposition (VD) algorithms can only be applied to cooperative tasks.

    \item Multi-agent Trust Region Learning (HATRPO \& HAPPO \citep{kuba2021trust}) can only be applied to cooperative tasks with shared critic models but seperately used actor models.
    \item COMA \citep{foerster2017counterfactual} can only be applied to tasks with discrete action spaces.
    \item MADDPG \citep{lowe2017multi}, and FACMAC \citep{peng2021facmac} can only be applied to tasks with continuous action spaces. 
\end{itemize}

There are additional cases that deviate from the goal of having a "one learning pipeline for all tasks." In other words, while the agent-level dataflow provides a tool for unifying algorithms in MARLlib, there is a lack of guidance on how to use this tool effectively. To address this, we introduce the compatibility test and auto-adaptation mechanism as an essential component of MARLlib, which aims to achieve the final unification of algorithms.

\subsection{Auto Adaption}

MARLlib's Auto Adaptation mechanism is a versatile feature that facilitates the selection of the optimal module based on the user's chosen algorithm and specific environments or tasks. This mechanism intelligently analyzes the algorithm's characteristics and the given environment, dynamically determining the most suitable module for the learning pipeline. By leveraging this adaptive capability, MARLlib ensures efficient and effective performance by utilizing the module that aligns with the algorithm's requirements and the intricacies of the task. This flexibility empowers users to seamlessly integrate diverse algorithms and tackle various tasks with enhanced adaptability, ultimately augmenting the overall effectiveness of MARLlib's learning pipeline.

	\begin{figure}[h!]
		\centering 
		
		\begin{subfigure}{1.0\textwidth}
			\includegraphics[width=\linewidth]{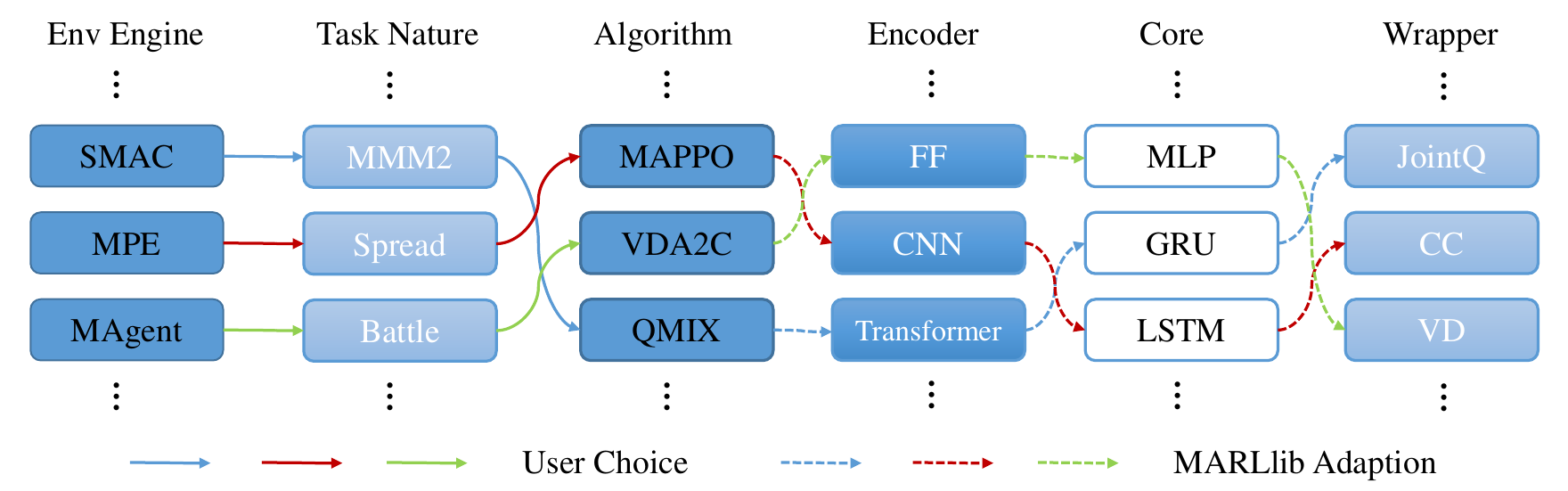}
		\end{subfigure}\hfil 
		\caption{The data flow begins with the user selecting a task and an algorithm to run against it. MARLlib then automatically adapts the module needed in the rest of the learning pipeline. Notably, the modularization aspect of MARLlib enables the automatic selection and combination of smaller module components, contributing to the overall adaptability and efficiency of the learning pipeline.}
		\label{fig:autoadaption}
	\end{figure}

\subsection{Compatiabilty Test}

The compatibility test function interface incorporates configuration aspects from four key components:
1) Training information,
2) Environment instance,
3) Model information, and
4) Stop condition.
These four parts collectively initiate a MARL process. Compatibility assessment begins by retrieving environment features such as agent number, agent name, and policy sharing restrictions. It then verifies the legality of user-defined policy sharing settings with respect to the given task. For example, the default option disables policy sharing for tasks with a large number of agents to conserve memory usage. Attempting to switch to a no-sharing policy strategy would trigger an error.

In addition to evaluating policy sharing compatibility on the task side, the algorithm side is also examined to ensure its suitability for the current tasks. For instance, if an algorithm is designed for discrete action spaces, tasks built for continuous control will be filtered out. If an algorithm requires training under a specific parameter sharing style, an error will be triggered if parameter sharing across agents is mandatory for the given task.

It is important to note that MARLlib categorizes different types of algorithms into distinct compatibility test suites. There are three compatibility tests in MARLlib, namely: independent learning (\texttt{run\_il}), centralized critic (\texttt{run\_cc}), and value decomposition (\texttt{run\_vd}). This organization allows algorithms within the same group to share more similarities, thereby reducing the complexity of assessing whether specific training requirements can be met based on the provided task/algorithm combination.

\subsection{Pipeline Unification on Agent-level Dataflow}

	\begin{figure}[h!]
		\centering 
		
		\begin{subfigure}{1.0\textwidth}
			\includegraphics[width=\linewidth]{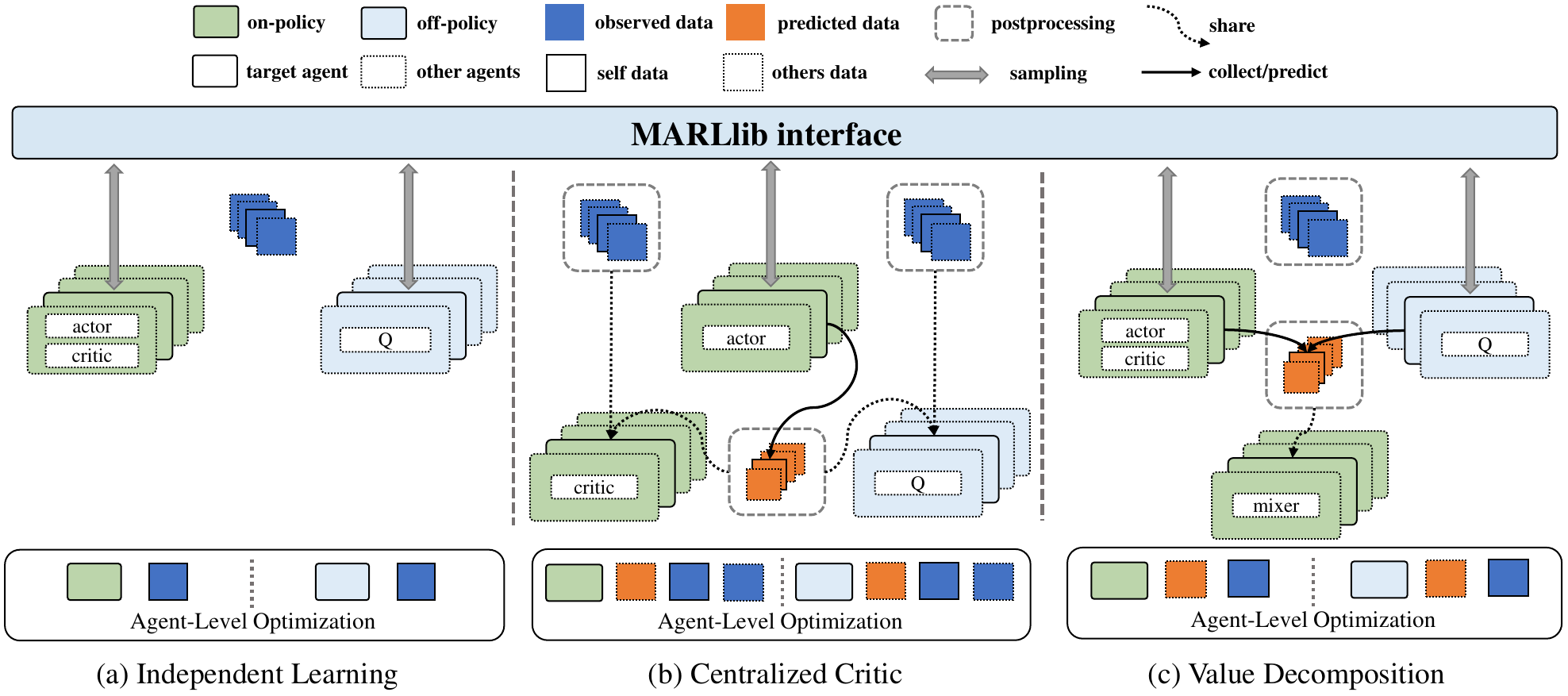}
		\end{subfigure}\hfil 
		\caption{The agent-level dataflow of MARLlib. The collection and sharing of observed and predicted data are in three distinct learning styles: independent learning, centralized critic, and value decomposition. 
		}
		\label{fig:algorithm}
	\end{figure}

MARLlib follows the Centralized Training Decentralized Execution (CTDE) framework to address multi-agent problems. In this framework, agents maintain their own policies for independent execution and optimization, while centralized information can be utilized to coordinate agents' update directions during the training phase.

Existing libraries typically split the whole learning pipeline into two stages: data sampling and model optimization. In the model optimization stage, all data sampled in the data sampling stage are available to make the training centralized. However, this approach couples the selection of proper data and the use of that data to optimize the model in the same stage. As a result, extending an algorithm to fit other task modes, such as both cooperative and competitive scenarios, becomes challenging and may require redesigning the entire learning pipeline.

To address this issue, MARLlib decomposes the original grouped dataflow into an agent-level distributed dataflow. Each agent in multi-agent training is treated as an independent unit during sample collection and optimization, but centralized information is shared among agents during the \texttt{postprocessing} phase to ensure equivalence. In this phase, agents share observed data (data sampled from the environment) and predicted data (actions taken by their policies or Q values) with others. All agents maintain individual data buffers to store their experiences and necessary information shared by other agents. Once the learning stage begins, no further information sharing is required among agents, and they can optimize themselves independently. This distribution of the dataflow to agents and complete decoupling of data sharing and optimization allows the same implementation to handle multiple task modes.

Furthermore, while all CTDE-based algorithms share a similar agent-level dataflow in general, they still have unique data processing logic. Inspired by EPyMARL \citep{papoudakis2021benchmarking}, the algorithms are further classified into independent learning, centralized critic, and value decomposition categories to enable module sharing and extensibility. Independent learning algorithms let agents learn independently; centralized critic algorithms optimize the critic with shared information, which then guides the optimization of decentralized actors; value decomposition algorithms learn a joint value function as well as its decomposition into individual value functions, which agents then employ to select actions during execution. Depending on their algorithmic properties, suitable data sharing strategies are implemented in the \texttt{postprocessing} phase, as illustrated in Figure \ref{fig:algorithm}.

In summary, the agent-level distributed dataflow in MARLlib unifies diverse learning paradigms while preserving the unique properties of all algorithms. This implementation approach allows a single pipeline to handle all task modes and remains equivalent to the original implementation.

\subsection{Dataflow equivalance}
    \label{equivalence}

In this section, we aim to substantiate the assertion that all multi-agent learning paradigms can be seamlessly transformed into amalgamated single-agent learning processes. Our goal is to demonstrate that the data used for optimizing the target entity remains unequivocally identical. To achieve this, we will utilize the well-established PyMALR and EPyMARL's learning framework, known for its efficacy. This framework comprises three essential components: sampling, centralized data collection, and one-step training. We will compare this approach with the distinct method employed by MARLlib, which is characterized by its sampling technique, sharing mechanism (post-processing), and agent-level training. The three different learning styles will be carefully examined and analyzed in the following discussion.

\begin{figure*}[!h]
		\centering 
		\includegraphics[width=\linewidth]{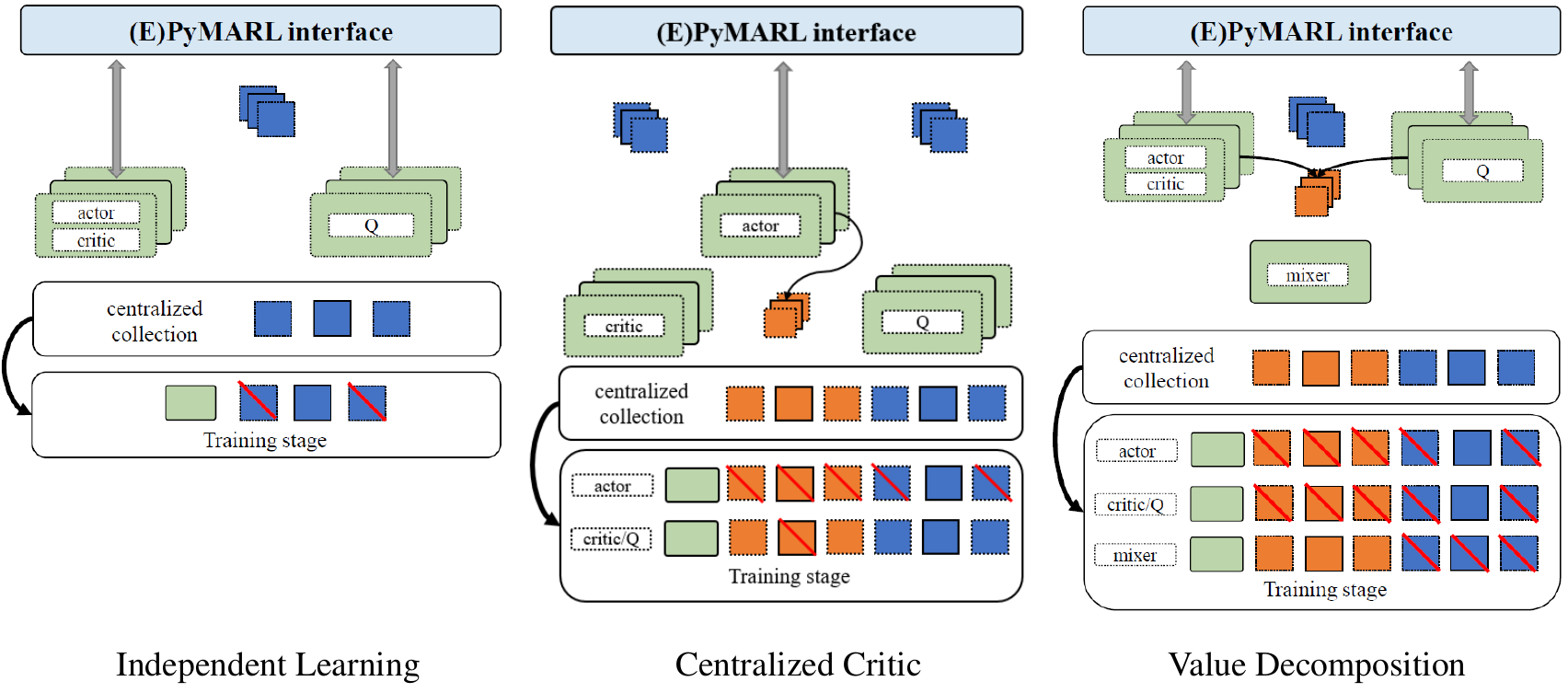}
		\caption{Learning pipelines of (E)PyMARL on independent learning, centralized critic, and value decomposition. Please note that for better illustration, only one agent is depicted in the training stage, which is not the real case for (E)PyMARL since all the agents should be trained simultaneously.}
		\label{fig:flow0}
   \end{figure*}

\begin{figure*}[!h]
		\centering 
		\includegraphics[width=\linewidth]{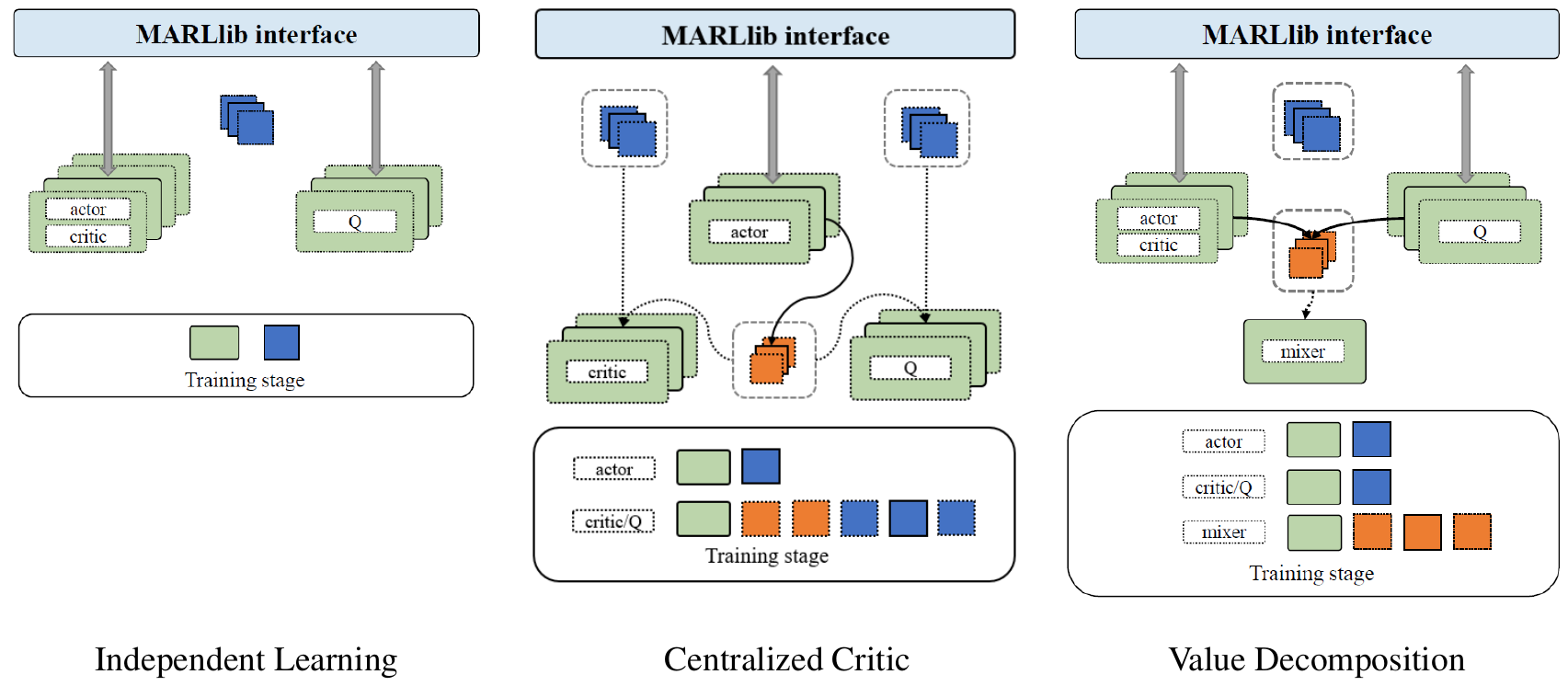}
		\caption{Learning pipelines of MARLlib on independent learning, centralized critic, and value decomposition.}
		\label{fig:flow1}
   \end{figure*}

{\bf Independent learning}: The left parts of Fig.~\ref{fig:flow0} and Fig.~\ref{fig:flow1} highlight the differences between (E)PyMARL's and MARLlib's independent learning pipelines. Before training, agents sampled data in group and put all data into one centralized data buffer. During the training stage, (E)PyMARL agents select the data from the centralized buffer, while MARLlib agents do not need to select since each agent maintains its individual buffer. In the end, same data is used to optimize the current agent's policy. As a result, two learning pipelines are equivalent.

    \label{sec:Centralized Critic}
{\bf Centralized Critic}: The middle parts of Fig.~\ref{fig:flow0} and Fig.~\ref{fig:flow1} illustrate the discrepancies between the centralized critic learning pipelines of (E)PyMARL and MARLlib. Simliar to independent learning, (E)PyMARL's pipeline involves feeding all collected data to the agent during the training stage. The agent then filters out unnecessary data to optimize the relevant parts, such as the critic/Q function that must ignore self-predicted actions. To the opposite, MARLlib's agent-level pipeline prepares all the required data for optimizing different model parts through postprocessing before the training stage. This makes the optimization process much easier to conduct and is independent of any specific learning style (e.g., same objective function for independent PPO and MAPPO). In the end, as shown in the training part, same data is used to optimize each agent, thus ensuring equivalence between (E)PyMARL's and MARLlib's learning pipeline.

{\bf Value Decomposition}: The differences between PyMARL's and MARLlib's value decomposition learning pipelines are illustrated in the right parts of Fig.~\ref{fig:flow0} and Fig.~\ref{fig:flow1}. Much similiar to centralized critic, in (E)PyMARL's learning pipeline, agents are fed with all data collected, pick the data they needed, and optimize different parts of their model. While in MARLlib's learning pipeline, all necessary data, including the Q/critic value are shared in the postprocessing stage. One agent can again simply optimize itself using all the data it maintains without considering other agents. We can also find that two learning pipelines use same data to update the policy. Therefore, the equivalence on the value decomposition learning style is established.

	\section{Extensibility}
	
The extensibility of MARLlib is guaranteed by its highly abstracted architecture, which consists of five major parts: configuration, training script, algorithm, model, and environments. Each part has corresponding APIs, methods, or instructions explained in the documentation, allowing for easy customization and extension. Here, we provide a brief description of how MARLlib's extensibility can be utilized in different aspects of MARL research:
	
	\begin{itemize}
		\item To extend one algorithm to tackle more task modes, such as transitioning from cooperative-only to mixed scenarios, one can focus on the script part provided under the \texttt{marl/algo/scripts} directory. For example, enabling an algorithm to handle additional task modes may involve reorganizing the policy mapping function in the script.
		\item To enable one algorithm to handle complex task data structures or partial observable settings, one can focus on the \texttt{marl/models} directory. For instance, for a well-known large-scale multi-agent task like Neural-MMO \citep{suarez2021neural}, which returns an observation containing both 3D observation and 1D state, adjustments can be made in this directory.
		\item To develop a completely new algorithm, one should focus on the algorithm part and build new functions based on existing basic algorithm implementations found under the \texttt{marl/algos/core} directory and \texttt{marl/utils} directory. For example, customizing the representation of global information for a new information-sharing mechanism can be done in \texttt{marl/algos/utils/centralized\_critic.py}.
		\item To quickly build a baseline for newly proposed multi-agent tasks that are not already included, attention should be given to the \texttt{envs} directory. By following the MARLlib agent-environment interface introduced in Section~\ref{sec:design} and Appendix~\ref{sec: task-side uni}, a new task can be easily incorporated, and all algorithms available in MARLLib can be evaluated on this new task.
	\end{itemize}

In conclusion, MARLlib's extensibility covers all aspects of conducting diverse MARL experiments, making it a powerful tool for MARL research. Additional examples can be found in the \texttt{examples} directory in the MARLlib repository.
    
	\section{Benchmarking}
	\label{sec:results}

In this section, we conducted a comprehensive evaluation of 17 algorithms on 23 tasks from five widely-used MARL testing environments, namely SMAC \citep{samvelyan2019starcraft}, MPE \citep{lowe2017multi}, GRF \citep{kurach2019google}, MAMuJoCo \citep{peng2021facmac}, and MAgent \citep{zheng2017magent}. We selected these environments for their popularity in MARL research and their diversity in task modes, observation shapes, additional information, action spaces, sparse or dense rewards, and homogeneous or heterogeneous agent types.

\begin{figure*}[h!]
		\centering 
		\includegraphics[width=\linewidth]{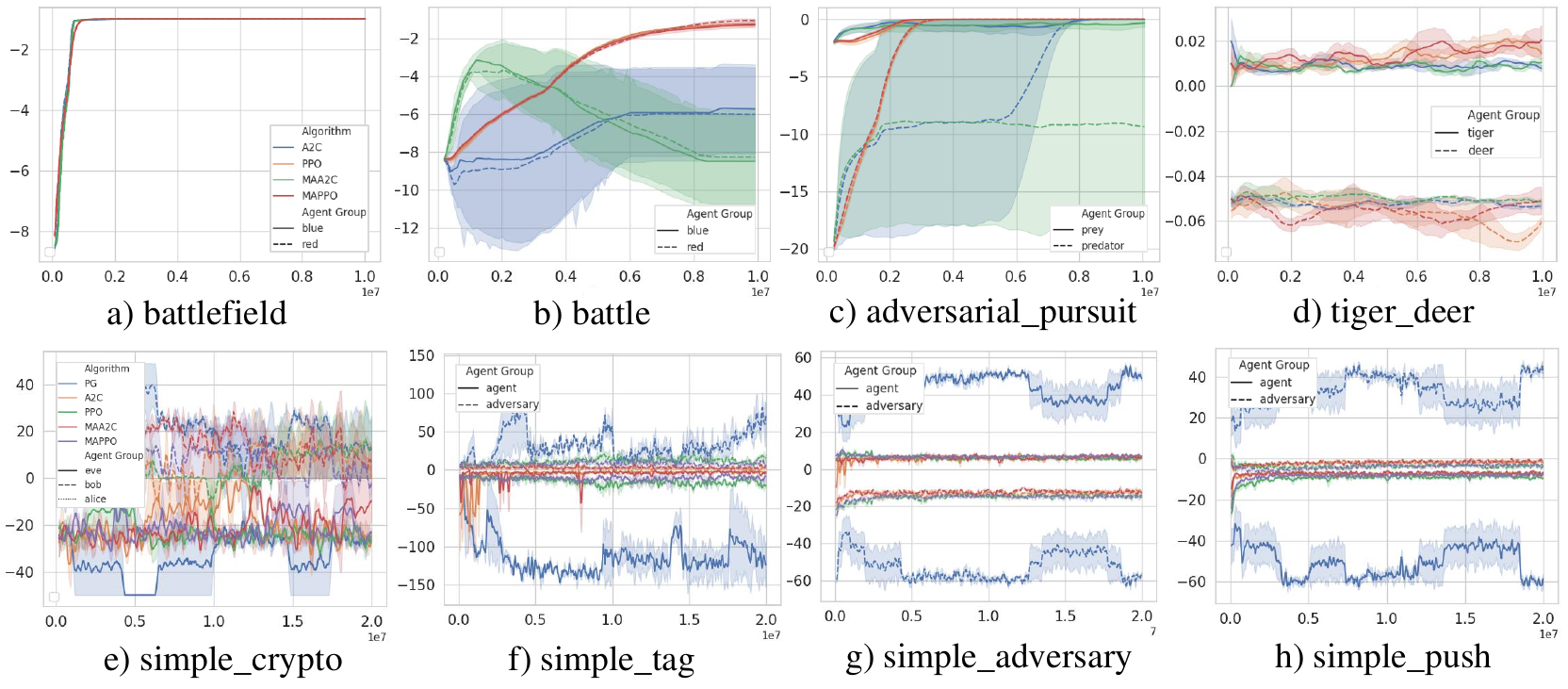}
		\caption{The figure displays the return curves for eight mixed scenarios where agents compete in groups, four from the MAgent environment (a-d) and four from the MPE environment (e-h). Each curve corresponds to a different agent group. The return curves of competing groups demonstrate a dynamic balance during the learning procedure, with the balance point varying based on both algorithms and tasks. For better visualization, the figure is zoomed in to highlight the details of the learning process.}
		\label{fig:mixed_algo}
	\end{figure*}

The evaluation involved running each algorithm on each task with four different random seeds, resulting in over one thousand experiments in total. We measured the mean return achieved by each algorithm across these experiments. The results of our experiments are presented in Table~\ref{table:algo_per} and Figure~\ref{fig:mixed_algo}. Based on these results, we were able to substantiate the quality of our implementation and provide insightful analysis.

    \begin{table*}[b!]
		\scriptsize
		\caption{
The table presents the algorithm performances (measured in return) for cooperative tasks, covering both discrete control tasks (MPE, SMAC, GRF) and continuous control tasks (MAMuJoCo). The evaluation includes four environment suites, with SMAC having two rows for each scenario. The first row (in italics) displays the performances reported by EPyMARL, while the second row shows the performances achieved using MARLlib. For the other environments, only MARLlib performances are included. Cells with a '-' indicate that no data was reported for that particular scenario. Dark cells highlight the top two performances in each scenario.
		}
		\vspace{-0pt}
		\begin{center}
			\resizebox{1.0\linewidth}{!}{
				\setlength{\tabcolsep}{0.5em}
				{\renewcommand{\arraystretch}{1.0}
					\begin{tabular}{ccccccc|cccc|cccc}
						\hline
						\multirow{2}{*}{Env}      & \multirow{2}{*}{Scenario}    & \multicolumn{5}{c|}{Independent Learning}                         & \multicolumn{4}{c|}{Centralized Critic}                             & \multicolumn{4}{c}{Value Decomposition} \\  
						&                              & IQL     & IPG     & IA2C    & ITRPO   & IPPO & MAA2C    & COMA    & MATRPO  & MAPPO   & VDN      & QMIX     & VDA2C   & VDPPO   \\ \hline
						\multirow{10}{*}{\rotatebox[origin=c]{90}{SMAC}}    & \multirow{2}{*}{2s\_vs\_1sc}                   & \it{16.72   }&\it{ -       }&\it{ 20.24   }&\it{ -       }&\it{ 20.24   }&\it{ 20.20    }&\it{ 11.04   }&\it{ -       }&\it{ 20.25        }&\it{ 18.04    }&\it{ 19.01    }&\it{ -       }&\it{ - }      \\
						&                                                & 16.09   & 20.07   & 20.07   & 20.16   & 20.18   & 20.09    & 10.32   & \cellcolor{gray!40} 20.23   & 20.21    & 16.3     & 17.25    & 15.61   & \cellcolor{gray!60} 20.24   \\ \cline{2-2} 
						& \multirow{2}{*}{3s5z}                          &\it{ 16.44   }&\it{ -       }&\it{ 18.56   }&\it{ -       }&\it{ 13.36   }&\it{ 19.95    }&\it{ 18.90   }&\it{ -       }&\it{ 19.91          }&\it{ 19.57    }&\it{ 19.66    }&\it{ -       }&\it{ -}       \\
						&                                                & 16.73   & 10.78   & 13.49   & 10.04   & 14.3    & 15.21    & 9.78    & 12.1    & \cellcolor{gray!60} 19.52   & \cellcolor{gray!40}19.38    & 19.32    & 8.58    & 13.15   \\ \cline{2-2} 
						& \multirow{2}{*}{MMM2}                          &\it{ 13.69   }&\it{ -       }&\it{ 10.70   }&\it{ -       }&\it{ 11.37   }&\it{ 10.37    }&\it{ 6.95    }&\it{ -       }&\it{ 17.78          }&\it{ 18.49    }&\it{ 18.40    }&\it{ -       }&\it{ -  }     \\
						&                                                & 12.08   & 9.21    & 10.17   & 8.04    & 10.37   & 16.08    & 6.7     & 7.62    & 16.86       & \cellcolor{gray!60} 19.31    & \cellcolor{gray!40} 18.34    & 2.72    & 9.31    \\ \cline{2-2} 
						& \multirow{2}{*}{3s\_vs\_5z}                    &\it{ 21.15   }&\it{ -       }&\it{ 4.42    }&\it{ -       }&\it{ 19.36   }&\it{ 6.68     }&\it{ 3.23    }&\it{ -       }&\it{ 18.17          }&\it{ 19.03    }&\it{ 16.04    }&\it{ -       }&\it{ -   }    \\ 
						&                                                & 16.78   & 5.6     & 10.79   & 3.39    & 7.95    & 12.14    & 4.79    & 13.32   & 17.24        & \cellcolor{gray!40} 18.55    & \cellcolor{gray!60} 19.84    & 9.6     & 14.61   \\ \cline{1-2}
						\multirow{3}{*}{\rotatebox[origin=c]{90}{MPE}}      & simple\_spread                                 & -197.61 & \cellcolor{gray!40} -63.83  & \cellcolor{gray!60} -63.16  & -78.16  & -65.74  & -63.37   & -71.64  & -77.63  & -66.26  & -190.5   & -189.27  & -190.66 & -213.99 \\
						& simple\_speaker\_listener                      & -44.07  & -261.65 & -29.06  & -50.17  & -38.29  & \cellcolor{gray!40} -27.76   & -67.6   & -44.01  & -34.41    & -35.26   & \cellcolor{gray!60} -25.68   & -54.37  & -64.61  \\
						& simple\_reference                              & -75.36  & -36.3   & -35.95  & -57.79  & -50.92  & \cellcolor{gray!40} -35.05   & -56.5   & -47.71  & -37.89     & -70.56   & \cellcolor{gray!60} -31.53   & -69.35  & -73.82  \\ \cline{1-2}
						\multirow{3}{*}{\rotatebox[origin=c]{90}{GRF}}      & pass\_and\_shoot                  & -0.17   & \cellcolor{gray!40} 0.6     & -0.03   & \cellcolor{gray!40} 0.6     & 0.5     & -0.02    & -0.01   & 0.48    & \cellcolor{gray!60} 0.74    & -0.06    & -0.24    & 0.05    & 0.01    \\
						& run\_pass\_and\_shoot             & -0.15   & \cellcolor{gray!60} 0.07    & -0.07   & -0.05   & -0.07   & -0.05    & -0.03   & \cellcolor{gray!40} -0.02   & -0.03     & -0.24    & -0.11    & -0.09   & -0.13   \\
						& 3\_vs\_1\_with\_keeper                         & 0.02    & 0.33    & 0.01    & \cellcolor{gray!40} 0.37    & 0.05    & 0        & 0.03    & 0.13    &\cellcolor{gray!60}  0.45       & -0.08    & -0.06    & 0       & 0       \\ \hline & & & & & & & & & & & & & &  \\
						&                                        & IPG     & IA2C    & IDDPG   & ITRPO   & IPPO    & MAA2C    & MADDPG   & MAPPO   & HAPPO   & FACMAC   & VDA2C    & VDPPO   &         \\  \cline{1-15}
						\multirow{5}{*}{\rotatebox[origin=c]{90}{MAMuJoCo}} & 2AgentAnt                                      & 143.22  & -268.02 & 44.60   & \cellcolor{gray!40} 527.10  & -153.46 & \cellcolor{gray!60} 730.8    & 18.53   &    -57.02      &  330.12   & -1224.6  & 449.19   & -98.74  &         \\
						& 2AgentHalfCheetah                              & -133.06 & -457.11 & -197.85 & \cellcolor{gray!60} 1652.49 & -644.89 & -493.3   & -313.95 &   -357.78 & \cellcolor{gray!40} 153.2 & -433.61  & -423     & -644.53 &         \\
						& 2AgentWalker                                   & 50.67   & 114.1   & 95.76   & \cellcolor{gray!60} 272.41  & 8.71    & 103.65   & 153.93  &    -4.12   & \cellcolor{gray!40}  164.45  & -7.88    &  125.49   & -3.76   &         \\
						& 4AgentAnt                                      &  \cellcolor{gray!40}  584.75  & 49.36   & -971.28 & \cellcolor{gray!60} 750.96  & -127.43 & -1005.30 & -419.93 &  149.1   &   151.85      & -457.68  & -338.21  & -164.72 &         \\
						& 6AgentHalfCheetah                              & -140.96 & -302.99 & -196.46 & \cellcolor{gray!60} 1492.24 & -653.78 & -257.76  & -207.49 &  -529.43 & \cellcolor{gray!40} 442.48 & -151.95  & -588.66  & -544.29 &         \\ \hline
					\end{tabular}
				}
			}
		\end{center}
		\label{table:algo_per}
	\end{table*}

To demonstrate the correctness of MARLlib, we conducted a comparison of its implementation on the SMAC environment with the performances reported by EPyMARL, while keeping the important hyper-parameters the same. The results obtained by EPyMARL involved 40 million steps for on-policy algorithms and four million steps for off-policy algorithms. In contrast, MARLlib consumed only half of these steps for training as we found it sufficient for convergence. Despite using fewer training steps, MARLlib matched most of the performances reported by EPyMARL, as depicted in Table~\ref{table:algo_per}.

When comparing the performance pairs available, MARLlib achieved similar results on 63\% of them (where the total reward difference is less than 1.0). Additionally, it outperformed EPyMARL on 25\% of the comparisons, and appeared slightly inferior on the remaining 12\%. Since each algorithm exhibited expected performances without relying on task-specific tricks, our implementation demonstrated generality and stability. These experimental results effectively substantiate the correctness of the MARLlib implementation.

Furthermore, the table also presents, for the first time, the performances of five algorithms on SMAC and MPE, twelve algorithms on GRF, and ten algorithms on MAMuJoCo, providing valuable reference points for the community.

\section{Future Development: Towards Real Application}

The potential for real-world applications of MARLlib in multi-agent reinforcement learning has garnered significant attention from researchers and practitioners. To further advance its utility and practicality, several key areas of future development warrant exploration.

\begin{itemize}
    \item Scalability and Efficiency: As MARLlib is extended to larger-scale scenarios, addressing scalability and efficiency becomes paramount. Future efforts should focus on optimizing algorithms and data structures within MARLlib to accommodate a higher number of agents, larger state and action spaces, and complex interactions. This will empower MARLlib to effectively address real-world challenges such as traffic management, swarm robotics, and large-scale industrial systems.
     \item Explainability and Interpretability: Enhancing the explainability and interpretability of MARLlib's decision-making processes is essential for its integration into real-world applications. Future endeavors should focus on developing techniques that enable visualizations of agent behaviors, insights into learned policies, and explanations of the underlying reasoning behind agent actions. These advancements will improve the transparency and understanding of MARLlib, particularly in domains that require interpretability, such as healthcare, finance, and autonomous systems.
     \item Robustness and Safety: Ensuring the robustness and safety of MARLlib in real-world applications is paramount. Addressing uncertainties and adversarial environments is crucial for its practical deployment. Future developments should concentrate on designing mechanisms that can handle partial observability, non-stationary environments, and communication failures among agents. By fortifying MARLlib's resilience, it can exhibit reliable and dependable behavior in real-world scenarios.
\end{itemize}

In conclusion, future development of MARLlib should prioritize scalability, explainability, and robustness regarding the integration with real-world systems. By focusing on these areas, MARLlib can transition from a research tool to a practical framework, enabling its application in diverse domains and propelling advancements in multi-agent reinforcement learning.

\bibliography{bibtex}

\end{document}